\documentclass{isprs} 
\usepackage{subfigure}
\usepackage{amssymb}
\usepackage{mathtools}
\usepackage{setspace}
\usepackage{geometry} 
\usepackage{epstopdf}
\usepackage[labelsep=period]{caption}  
\usepackage[british]{babel} 
\usepackage[hang]{footmisc}
\usepackage{graphicx}
\usepackage[breaklinks=true, hidelinks]{hyperref}

\begin{document}

\title{Synthesizing hyperspectral Data using generative Models to train spectral Unmixing Methods for low-cost Crop Residue Cover Mapping}
\date{}

\author{
Ege Artan\textsuperscript{1,3}\thanks{Corresponding author} , Shuo Chen\textsuperscript{2,3}, Andrew Peng\textsuperscript{2,3}, Sammuel Aldrich Karya\textsuperscript{2,3}, Kyaw Thiha\textsuperscript{1,3} 
}

\address{
	\textsuperscript{1 } University of Toronto Scarborough, Univ. of Toronto, 1265 Military Trail, Toronto, ON M1C 1A4, \\ Canada - (ege.artan, kyaw.thiha)@mail.utoronto.ca \\
	\textsuperscript{2 } Faculty of Arts and Science, University of Toronto, 100 St. George Street, Toronto, ON M5S 3G3, \\ Canada - (michaeldavid.chen, andrewpeng.peng, sam.karya)@mail.utoronto.ca \\
	\textsuperscript{3 } Space Systems, University of Toronto Aerospace Team, 10 King's College Road, \\ B740 Sandford Fleming Building, Toronto, ON M5S 3G8, Canada \\
}

\abstract{
The spectral range of the Field Imaging Nanosatellite for Crop residue Hyperspectral Mapping (FINCH), developed by the University of Toronto Aerospace Team's Space System Division, poses significant challenges for hyperspectral unmixing to determine crop residue cover fractional abundances. The severe lack of standard indices necessitates the use of complex, data-driven unmixing models. Complex unmixing models require dense, well-distributed manifolds, whereas ground-truth datasets are sparse and expensive. To better leverage the information content in existing ground-truth datasets, this study presents a framework that decouples interpolation and abundance-mapping estimation via an intermediary conditional data generator, contrasting with traditional single-model unmixing pipelines. This decoupling enables the inclusion of physical prior knowledge about spectra, which is not natively accessible to unmixing models, thereby artificially expanding the dataset to serve as a Monte Carlo approximation of the population risk. To test this hypothesis, we have proposed two generator models: a 1D Conditional U-Net with Conformer Layers (GD-Streamline) and a Dual-Path Transformer Conditional Variational AutoEncoder (TCVAE), and two unmixing models: Multi-Layered Perceptrons (MLP) and Fourier Neural Operators (FNO). The results of the study indicate a need for rigorous integration and introduction of spectral priors within the scope of the proposed decoupling framework to prevent domain and mode collapse and hallucination by the generator models; otherwise, this leads to falsely approximated data manifolds, ultimately resulting in unsatisfactory and out-of-distribution unmixing performance.
}


\maketitle

\section{Introduction}\label{sec:Introduction}
\sloppy

The University of Toronto Aerospace Team's (UTAT) Space Systems Division is developing the Field Imaging Nanosatellite for Crop residue Hyperspectral mapping (FINCH) CubeSat, a mission focusing on crop residue cover (CRC) mapping using the 900--1700 nm (SWIR1) spectral range. As such, a core objective for FINCH's Science team is to develop a hyperspectral unmixing pipeline for the given spectral range and to determine the high-level engineering requirements for FINCH's payload and bus teams. The spectral range available to FINCH rules out the use of well-known indices, as there are no significant spectral features. As a result, FINCH's Science team has been developing statistical, machine, and deep learning-based models, given their growing promise for CRC estimation ~\cite{yang2025}. 

However, developing such data-driven unmixing models requires dense, high-quality, and well-distributed datasets that are similar to the data that would be downlinked from FINCH in terms of atmospheric effects. To simulate the effects caused by the atmosphere, the team has increasingly focused on developing an atmospheric modeling and inversion pipeline using the ground-truth dataset. Additionally, such datasets must have a dense, wide distribution over abundance (and consequently spectral) space, so that the data-driven model has readily accessible information for proper training. However, a gap remains in the density and distribution of the current ground-truth dataset used by the Science team ~\cite{dennison2019}, and this gap similarly persists in various hyperspectral remote sensing applications ~\cite{yang2025}. The low density and improper distribution of the pre-existing datasets used by the Science team become a roadblock for some high-capacity data-driven unmixing models.

 Addressing this data gap traditionally requires either developing highly data-efficient hyperspectral unmixing algorithms that better exploit the information available in the datasets or conducting extensive, logistically complex, and financially expensive ground-truth data-collection missions. We propose a new pipeline as a solution, contingent on the system's physics and requiring complex hyperspectral unmixing algorithms, but lacking expansive, well-distributed datasets. 

The described solution aims to increase the overall efficiency of the hyperspectral unmixing pipeline by decoupling spectral interpolation and spectra-to-abundance mapping into two separate models: a data generator trained on the ground-truth dataset to synthesize spectra, and an unmixing model trained on the synthesized spectra. We propose that, under certain conditions that also apply to the FINCH mission, such decoupling can, in theory, make the information content in ground-truth datasets much more readily accessible to unmixing models.

Following previous architectural screenings of various models and frameworks, the study has refined its focus to two architectures: a Gaussian Diffusion model (specifically, a Denoising Diffusion Implicit Model) based on a 1D Conditional U-Net with Conformer Layers, and a Dual-Path Transformer Conditional Variational AutoEncoder. This study uses a larger spectral range that is not available to FINCH (400--2490 nm) for the training of the generator model, to make use of a higher amount of information available in the wider spectral range, to resolve specific, yet not very significant, spectral features that are available in FINCH's spectral range, 900--1700 nm, SWIR1.

\section{Theory}\label{sec:Theory}

In the theory section, we formalize and provide reasoning for the proposed decoupling of interpolation and spectra-to-abundance mapping within the proposed pipeline. We hypothesize that by isolating the two tasks of interpolation and spectrum-to-abundance mapping, additional spectral information and physical prior knowledge can be used, thereby aiding learning of the physical interactions among spectral bands and the underlying constraints of the hyperspectral unmixing problem. Furthermore, we address, at the end of this section, the information-theoretic implications of the decoupled pipeline, specifically within the context of the Data Processing Inequality.

\subsection{Problem Definition and Proposed Solution}\label{sec:Problem Definition and Proposed Solution}

The objective of any unmixing model $U$, as presented in Equation \ref{eq:unmixer_def}, which maps any spectrum $s$ to abundance $a$, is to minimize the population risk, Equation \ref{eq:population_risk}, where $\mathcal{M}(s, a)$ is an unknown joint distribution, a manifold of spectra and abundance combinations, $\Lambda_1$, $\Lambda_2$ are the dimensions of spectrum and abundance vectors respectively, and $L$ is a loss function that defines the population risk, as shown in Equation \ref{eq:unmixer_def}. Given the FINCH mission problem definition, the abundance vector is a 3-dimensional vector of green vegetation, non-photosynthetic vegetation (i.e., crop residue), and soil fractional cover abundances.
\begin{equation}\label{eq:unmixer_def}
    U: s \rightarrow a \hspace{0.2cm} | \hspace{0.2cm} s \in \mathbb{R}^{\Lambda_1}, a \in \mathbb{R}^{\Lambda_2}
\end{equation}
\begin{equation}\label{eq:population_risk}
    R(U) = E_{(s,a) \sim \mathcal{M}}[L(U(s),a)] = \int_{\mathcal{S}\times\mathcal{A}}L(U(s), a) \ d\mathcal{M}(s,a)
\end{equation}
Given the actual physical limitations, that one cannot acquire the true manifold of the abundance and spectra $\mathcal{M}$, the population risk integral is consequently approximated by a dataset manifold of $n$ spectra, $\mathcal{M}_n = \{(s_i, a_i)\}^{n}_{i=1}$, to the empirical risk minimization:
\begin{equation}\label{eq:empr_risk_gt}
    \widehat{R}_n(U) = \frac{1}{n} \sum^{n}_{i=1}(L(U(s_i),a_i))
\end{equation}
Assuming that the loss function $L$ is bounded by an arbitrary constant $M$ while also being $\mu-$Lipschitz continuous, due to the inherent physics of the system bounding both abundances and spectra, Statistical Learning Theory establishes the gap between population and empirical risk ~\cite{mohri2018}. The established gap with probability $1-\delta$ is:
\begin{equation}\label{eq:pop_empr_gap}
    R(U) \leq \widehat{R}_n(U) + 2\mu \mathfrak{R}_{n}(U) + M\sqrt{\frac{\ln(1/\delta)}{2n}}
\end{equation}
where the $\mathfrak{R}_{n}(U)$ represents the Rademacher complexity of the class of model $U$ on dataset $\mathcal{M}_n$. Under the assumption that the unmixing model $U$ is well-trained to minimize $\widehat{R}_n(U)$, a considerably small $n$ (the size of the dataset) combined with a high-complexity model class that $U$ is in (to facilitate the learning of complex inter-band interactions) causes the Rademacher complexity (2\textsuperscript{nd} term) and 3\textsuperscript{rd} terms dominate over the empirical risk minimization ($\widehat{R}_n(U)$); which results in considerably high bound on $R(U)$.

Therefore, if we wish to approximate the population risk using a ground-truth dataset, $\mathcal{M}_n$, the approximation will always be bound by (2\textsuperscript{nd} term) and 3\textsuperscript{rd} terms on the RHS of Equation \ref{eq:pop_empr_gap}. As a solution, we propose the training of a generative model, $G$, which we assume to approximate the manifold $\widehat{\mathcal{M}}_k = \{(s_i^{gen}, a_i^{gen})\}^{k}_{i=1}\sim \mathcal{M}$ with $k$ many spectra generated, which will result in empirical risk minimization $\widehat{R}_k(U)$ as we present in Equation \ref{eq:empr_risk_gen}. 
\begin{equation}\label{eq:empr_risk_gen}
    \widehat{R}_k(U) = \frac{1}{k} \sum^{k}_{i=1}(L(U(s_i^{gen}),a_i^{gen}))
\end{equation}
Under the condition of a sufficiently large generated dataset ($k\gg n$), evaluating empirical risk $\widehat{R}_k(U)$ over $\widehat{\mathcal{M}}_k$ acts as a Monte Carlo approximation of the population risk ~\cite{caflisch1998}. To establish a strict upper bound on population risk, we must take into account the error introduced by a generator $G$, $\epsilon_{gen} \equiv ||\mathcal{M}-\widehat{\mathcal{M}}_k||$, and the unmixing model's approximation error, $\epsilon_{app}(U)$, which results in the following intermediary bound, with the empirical risk minimization for the generated dataset $R_{gen}(U)$:
\begin{equation}\label{eq:intr_bound}
    R(U) \leq R_{gen}(U) + \epsilon_{gen} + \epsilon_{app}(U)
\end{equation}

Following Equation \ref{eq:intr_bound}, substituting the bound on $R(U)$ using Equation \ref{eq:pop_empr_gap}, within the scope of a generated dataset with $k$ spectra, to the term $R_{gen}(U)$ yields the decoupling gap:
\begin{equation}\label{eq:decoupling_gap}
    R(U) \leq \widehat{R}_k(U) + 2\mu \mathfrak{R}_{k}(U) + M\sqrt{\frac{\ln(1/\delta)}{2k}} + \epsilon_{gen} + \epsilon_{app}(U) 
\end{equation}

Equation \ref{eq:decoupling_gap} implies that with an arbitrarily large generated dataset ($k\gg n$), the Rademacher complexity on the generated dataset $\widehat{\mathcal{M}}_k$ (2\textsuperscript{nd}) and confidence interval (3\textsuperscript{rd}) terms approach zero asymptotically. As a result of this asymptotic behavior at sufficiently large generated datasets, the population risk bound is dominated by the empirical risk on generated data (1\textsuperscript{st}), the generator's bias (4\textsuperscript{th}), and the unmixer's bias (5\textsuperscript{th}).

Given a standard, unmixing model only, supervised learning setup, with a small ground-truth dataset, a highly complex class of unmixing model $U$, trained to minimize the approximation error $\epsilon_{app}(U)$, the Rademacher complexity and confidence interval terms in Equation \ref{eq:empr_risk_gt} are susceptible to provide a high bound, resulting in poor generalization. However, the decoupling gap, presented in Equation \ref{eq:decoupling_gap}, with a sufficiently low-bias generator $G$, $\epsilon_{gen}$, generating a sufficiently large enough dataset ($k\gg n$), a highly-complex unmixing model class $U$ (used to capture complex physics of endmember mixing) can be trained to minimize $\widehat{R}_k(U)$, the empirical risk on the generated dataset. 

This result, in theory, presents a superior solution to a single unmixing model $U$, under the conditions that the model $U$ is highly complex, the generated dataset has many more samples than the ground-truth dataset ($k\gg n$), and a sufficiently low bias generator $G$ has been trained on the ground-truth dataset. In the FINCH mission, we are bound to use a relatively complex unmixing model class (due to the lack of indices and major spectral features in SWIR1) to achieve low approximation error, and we do not have a pre-existing expansive ground-truth dataset. Therefore, the proposed decoupling framework, under the conditions stated above, provides a mathematically grounded means of enabling complex unmixing models to reach their theoretical capacity by more efficiently processing the information in $\mathcal{M}_n$.

\subsection{Application on Manifold Hypothesis}\label{sec:Application on Manifold Hypothesis}

The Manifold Hypothesis implies that the spectra that exist in $\mathbb{R}^{210}$ of the ground-truth dataset used by the Science team, lie on a topological manifold $\mathcal{M}_n$ which has a much lower intrinsic dimension, due to physical constraints that limit the spectra themselves. When the universal approximators ($U$, unmixing models) are trained on this sparse ground-truth data, $U$ becomes highly unconstrained on the empty space that is $\mathbb{R}^{210}\setminus\mathcal{M}_n$, which leads to out-of-distribution instability. However, when training a generator model $G$, we can impose more constraints in its training, compared to $U$, since it trains in the spectral domain, such as the Spectral Angle (SAM), Total Variation (TV), and Mean Squared Error (MSE) on the Fast Fourier Transform (FFT). This results in the more efficient use of the information present in the ground-truth dataset, which is not directly accessible to $U$. As a result, local coverage around $\mathcal{M}_n$ can be increased within the scope of generator's inductive bias using physical prior information that can be introduced during the training of $G$ (which is inaccessible during the training of $U$, effectively reducing the empty space $\mathbb{R}^{210}\setminus\mathcal{M}_n$. 

\subsection{Effect on Lipschitz Continuity of Unmixing Models}\label{sec:Effect on Lipschitz Continuity of Unmixing Models}

A sufficiently robust and accurate unmixing function/model should exhibit Lipschitz continuity with a relatively small $K$, such that small perturbations on spectra should cause bounded, small variations on the predicted abundances, such that for all spectra $s_1, s_2 \in \mathcal{S}$:
\begin{equation}\label{eq:lipschitz}
    d_{\mathcal{A}}(U(s_1), U(s_2)) \leq Kd_{\mathcal{S}}(s_1, s_2)
\end{equation}
When data is sparse ($n\ll k$), the constant $K$ is often required to be high to account for a badly interpolated neighborhood around $s_1$ and $s_2$, due to possible high-frequency oscillations and poor generalization. By populating the manifold with generated samples, a form of Vicinal Risk Minimization that employs spectral priors is applied, which unlike augmentation through methods like broad-band or small-scale noise addition, that could push the model into the nonphysical regions of $\mathbb{R}^{210}$, contains the neighborhood of the ground-truth data points in a much more physically contained region. We expect this population of the manifold to reduce high-frequency oscillations during $ U$'s interpolations, thereby reducing $K$.

\subsection{The Role of Inductive Biases and the Data Processing Inequality}\label{sec:The Role of Inductive Biases and the Data Processing Inequality}

We note that any model or algorithm that processes data is bound by the Data Processing Inequality (DPI), which applies to the proposed decoupling by imposing an upper bound on the mutual information between a learned representation of $G$ and the abundance labels, $M_n$, while training a generator model $G$, which acts as an information content bottleneck. However, it must be acknowledged that DPI explicitly applies to the information content available, not necessarily to its accessibility or representational utility. The decoupled pipeline allows the introduction of several spectral priors, such as the SAM, TV, and FFT. Such spectral priors cannot be directly applied to the predictions of unmixing models and are not always accessible to them through architectural implementations, since, without knowing the exact mixing model of the system involved, a spectrum cannot be reconstructed to apply them. 

As a result, although DPI dictates that some information is lost, the decoupling of the unmixing pipeline allows us to consider that the inductive biases introduced by spectral priors may significantly enhance representational efficiency by indirectly generating samples.

\subsection{Summary of Assumptions and Claims}\label{sec:Summary of Assumptions and Claims}

It is imperative to restate the assumptions of the proposed decoupling of interpolation and mapping properties of an unmixing model $U$, into a generative model $G$ and an unmixing model $U$. Given the assumptions discussed in Section \ref{sec:Problem Definition and Proposed Solution}, the validity of the proposed decoupling requires the following: 
\begin{enumerate}
    \item The ground-dataset is insufficient to fully train an unmixing model $U$, with a high capacity, but is sufficient to train a generative model $G$, which can approximate the structure of the underlying manifold, through the use of additional physical priors that can be introduced during the training of the generative model $G$, as stated in Section \ref{sec:Application on Manifold Hypothesis}.
    \item The unmixing model $U$ intended to be trained is highly complex due to the complex interactions of the underlying physics, or the lack of easily identifiable spectral features.
    \item The generator $G$ is assumed to have been trained sufficiently such that it can approximate the underlying manifold, does not suffer from model and domain collapse, and does not have a high bias.
\end{enumerate}
Given the assumed conditions, we expect the unmixing model $U$ trained in the scope of such decoupling to:
\begin{enumerate}
    \item Make use of the additional spectral priors such as SAM and TV introduced during the training of $G$, resulting in more efficient use of the information present in the ground-truth dataset.
    \item In the majority of neighborhoods around the ground-truth data points for the unmixing model $U$, there are more well-behaved Lipschitz constants ($K$).
    \item Overall increase in the unmixing accuracies of the trained model $U$.
\end{enumerate}
In the later sections of this study, the main center of exploration becomes: ``to what extent does the proposed decoupling framework benefit the unmixing pipeline of FINCH?"

\section{Generative Model Definitions}\label{sec:Generative Model Definitions}

While initial conceptualizations and explorations for the generative model $G$ included standard Conditional Convolutional AutoEncoders (C-CVAE), Conditional Wasserstein Generative Adversarial Networks (C-WGAN), and Conditional Diffusion Transformers (C-DiT), the preliminary testing provided several challenges. The C-CVAE models were prone to mode collapse, C-WGANs had severe convergence difficulties, and C-DiT models suffered from vanishing gradients due to their deep architectures. Consequently, the preliminary architectures either underwent design iterations or were discontinued due to the higher technical demands posed by conditional generative models. C-CVAE architectures were evolved into a novel Dual-Path Transformer Conditional Variational AutoEncoder architecture to mitigate over-smoothing and mode collapse; C-DiT models were reformed into a Gaussian Diffusion model (DDIM) based on a 1D Conditional U-Net with Conformer Layers to mitigate vanishing gradients and denoising artifacts. The models defined in this section were selected for their high theoretical capacity and flexibility to include physical priors to better approximate the underlying data manifold given the constraints detailed in Section \ref{sec:Theory}.

\subsection{U-Net-based Gaussian Diffusion}\label{sec:Gaussian Diffusion}

\begin{figure}[htp]
    \centering
    \includegraphics[width=\columnwidth]{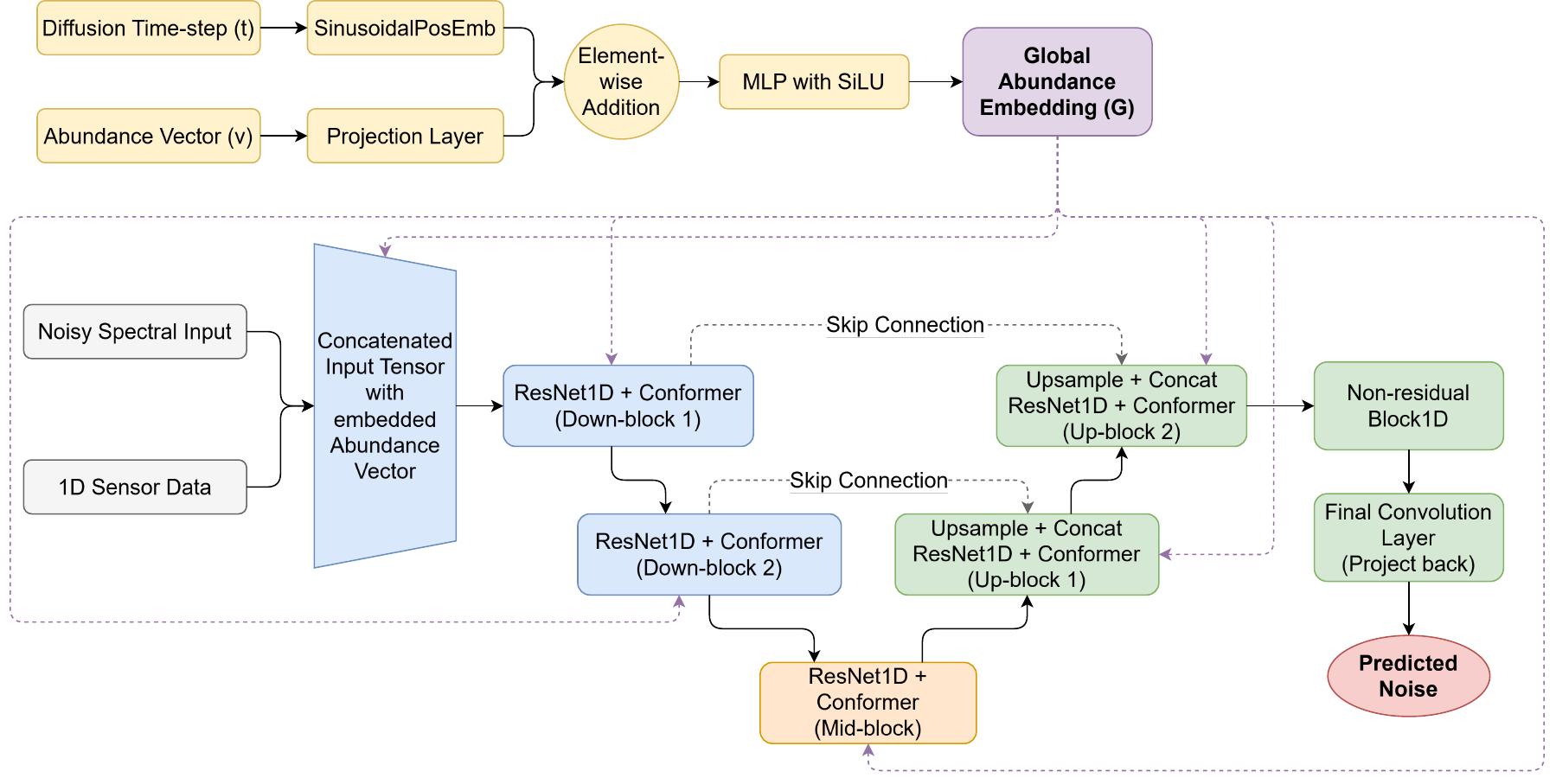}
    \caption{Architectural Diagram of GD-Streamline}
    \label{fig:gd-streamline}
\end{figure} 

The Gaussian Diffusion, selected for its high capacity and mode coverage, employs a 1D Conditional U-Net we adapted from Matcha-TTS ~\cite{mehta2024} for fixed-length spectral signatures as its backbone. We modified the architecture by replacing frame-wise phoneme encoding with global abundance conditioning, using Conformer blocks instead of standard Transformers, and implementing Mish activations instead of Snake activations, providing semantic guidance for spectral unmixing rather than temporal alignment.

As presented in Figure \ref{fig:gd-streamline}, the network conditioning integrates, via element-wise addition, the scalar diffusion time-step and the abundance vector after sinusoidal position embedding and projecting, respectively. The fused conditions are then forward propagated through an MLP with SiLU activations to yield a unified global embedding that modulates the U-Net feature maps.

The U-Net processes the input tensor constructed by concatenating the 1D sensor input (spectra) and the noisy spectral input. The architecture employs a symmetric hierarchy of down-blocks, mid-blocks, and up-blocks, connected by residual skip connections. Each layer employs ResNet1D blocks to integrate global abundance embeddings and uses Conformer layers to capture global features based on abundances and diffusion time steps.

The model denoises using a deterministic DDIM logic, which is amplified by classifier-free guidance during sampling and inference. Training minimizes a composite loss function that combines the standard L2 loss for noise prediction with physical priors on reconstructed spectra: spectral angle (SAM), L1 loss on Fast Fourier Transform (FFT) magnitudes, and Total Variation (TV). During training, the conditioning is occasionally dropped, enabling the model to learn to produce an unconditional output.

Ablation studies on architectural hyperparameters revealed that stacking multiple Conformer blocks within each ResNet1D block led to vanishing gradients, weakening visual heuristic performance during training. As such, restricting the architectures to employ a single Conformer block per ResNet1D module maximized the residual-to-computational-layer ratio, thereby mitigating gradient attenuation and improving convergence. Additionally, the number of ResNet1D blocks per level without the introduction of frequent residual connections hindered signal propagation and the learning of fine-grained features. As a result, the finalized architecture, employing one ResNet1D block per layer and one Conformer layer for each ResNet1D block, was named ``GD-Streamline".

\subsection{Dual-Path Transformer Conditional Variational AutoEncoder}\label{sec:TCVAE}

\begin{figure}[htp]
    \centering
    \includegraphics[width=\columnwidth]{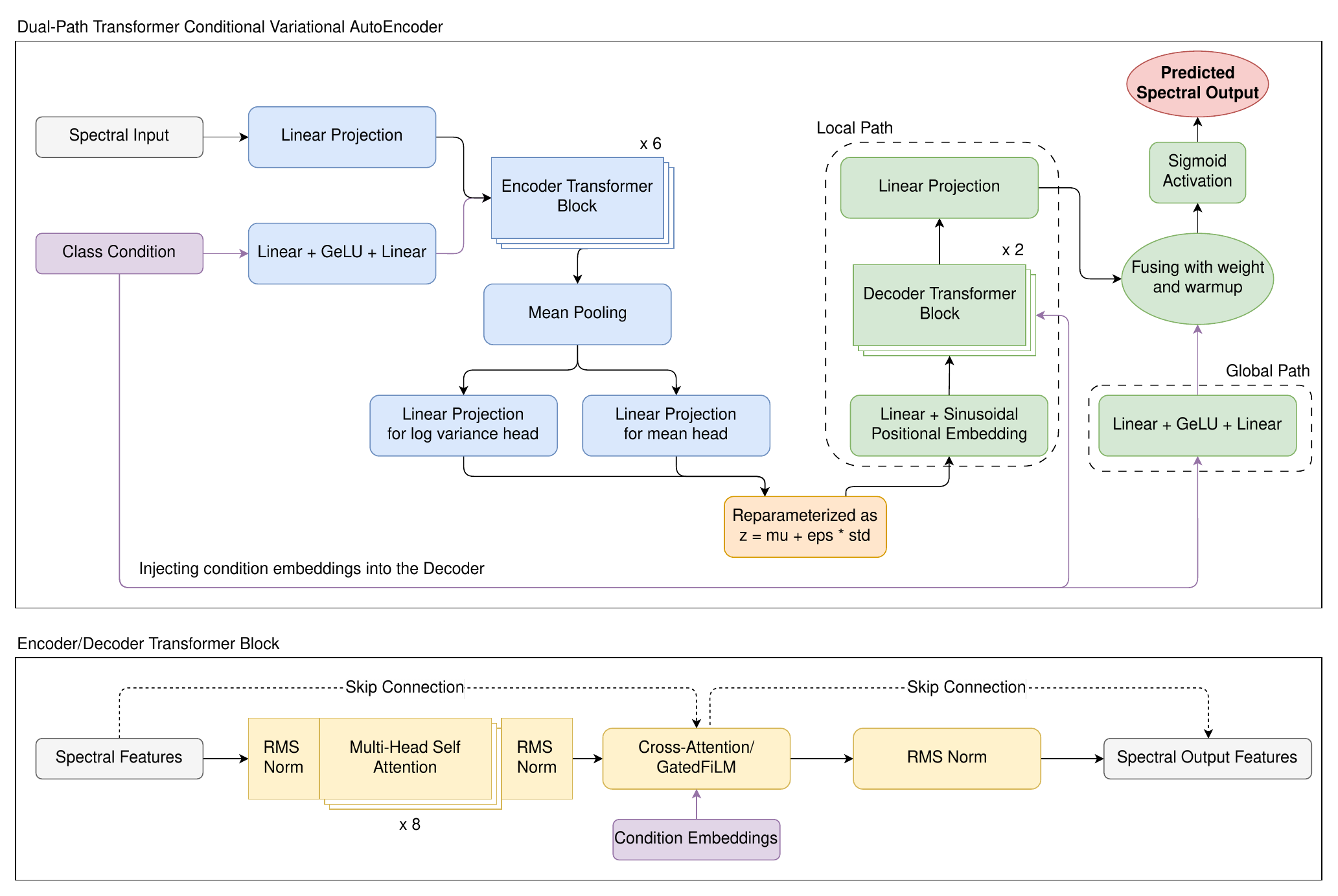}
    \caption{Architecture of Dual-Path TCVAE.}
    \label{fig:tcvae}
\end{figure} 

Dual-Path Transformer Conditional Variational AutoEncoder (Dual-Path TCVAE, referred to as TCVAE throughout this study) is the main AutoEncoder-based architecture we propose. Our earlier empirical results show a tendency for the Transformer and Conditional-VAE architectures to over-smooth the output, as the decoder relies entirely on the conditioning values for green vegetation, non-photosynthetic vegetation, and soil. To solve this, we propose a dual-path decoder: the global path uses condition values, and the local path uses only latent variables.

Specifically, our encoder consists of an MLP projection for class conditions and spectrum tokens, which are then fed into 6 Transformer blocks, followed by pooling and projection to the Gaussian distribution. Each Transformer block consists of a standard multi-head attention module with RMS Norm for feature extraction, followed by cross-attention with condition embeddings, and a final latent extraction using an MLP head.  Our decoder consists of two branches: a global path and a local path. The global path is a 2-layer MLP that processes the condition embedding. The local path consists of a linear latent projection, followed by a sinusoidal projection, then 2 decoder transformer blocks, and finally another learned linear projection. We fuse these global \& local spectra with learnable weights and a configurable warm-up scale. The architecture details can be found in Figure \ref{fig:tcvae}. 

\section{Test Metric Definitions}\label{sec:Test Metric Definitions}

\subsection{Nearest Neighbor}\label{sec:Nearest Neighbor}

We present a metric that quantifies how well the generated and ground-truth samples spread across each other's distributions, given the absence of pre-existing metrics for quantifying manifold approximation. The metric uses the nearest neighbors (defined by a distance metric) of a curated subset of ground-truth samples to generate samples. The dataset used for validation and testing during the generator's training was split into two distinct subsets, with the abundances of one subset used to generate spectra as generation conditions. The spectra were combined into an adjacency matrix with a directed 1-nearest-neighbor mapping to determine which spectra in a column had which spectra in a row as their nearest neighbors. Such an adjacency matrix allows the analysis of approximate distributions of the generated spectra within the ground-truth dataset by creating a network of nearest-neighbor relations, enabling the quantification of generation bias and variance. Two distance metrics, Euclidean and spectral angle, were used to identify the closest neighbors of each point, enabling detailed quantification of spectral amplitude and shape similarity, respectively. 

As a result, the number of encodings of $\gamma_{11}$ (top-left) and $\gamma_{22}$ (bottom-right) quadrants represents which distributions are more tightly clustered, with $\epsilon \coloneqq \gamma_{11}-\gamma_{22}$ representing the discrepancy. A high absolute value of the $\epsilon$ metric relative to the total number of spectra used can also indicate a mode collapse, in which the model generates highly similar spectra that are more tightly clustered. Additionally, as a consequence of the adjacency matrix, the number of encodings in $\gamma_{12}$ (top-right) quadrant represents the precision of generated spectra, i.e., how closely similar the generated spectra are to another ground-truth data. Consequently, the number of encodings in $\gamma_{21}$ (bottom-left) quadrant represents the recall of the generator. Thus, a positive $\zeta \coloneqq \gamma_{12}-\gamma_{21}$ represents a mode collapse given that the generated spectra collapse to a narrow physical space. In contrast, a negative $\zeta$ represents mode invention, meaning that the generator is hallucinating. Finally, a $\delta$ metric as constructed in Equation \ref{eq:metric_delta} represents the domain gap of the model, with $\delta \approx 1$ representing the ideal generator, and $\delta \gg 1$ representing a severe model gap in some capacity. 

\begin{equation}\label{eq:metric_delta}
    \delta \coloneqq \frac{\gamma_{11}+\gamma_{22}}{\gamma_{12}+\gamma_{21}}
\end{equation}

Thus, the nearest-neighbor definitions presented above are used as metrics to quantify how well the manifold of generated samples, $\widehat{M}_k$, approximates the manifold of the ground-truth dataset, $M_n$. Consequently, the metrics enable the interpretation of the scale of $\epsilon_{gen}$, the generator $ G$'s bias. All testing using nearest-neighbor metrics was conducted over the entire available spectral range of the ground-truth dataset, 400--2490 nm, to comprehensively quantify $\epsilon_{gen}$ across the full spectral range in which $G$ was trained.

\subsection{Hyperspectral Unmixing Models}\label{sec:Hyperspectral Unmixing Models}

We proposed several unmixing models for initial screening using the ground-truth dataset. Based on the ground-truth dataset, we filtered the models and selected two to test on the generated datasets to assess the validity of the decoupling proposal. We primarily quantified the models' accuracies using the mean $R^2$ (also called total $R^2$) of predicted versus actual abundance across the three endmember classes. This initial screening was conducted in the full spectral range available: 400-2490 nm.

\subsubsection{Multi-Layered Perceptron:}\label{sec:Multi-Layered Perceptron}

As a baseline, we use a fully-connected MLP that directly maps the 210-band vector to the three abundances. We have implemented two MLP architectures, where the first, MLP-S, has two hidden layers: the first with 64 neurons and the second with 32 neurons (each followed by ReLU activations). This yields a lightweight model ($\sim$16k parameters) compared to the CNN and FNO (which have millions of parameters). The second model, MLP-M, implemented four layers of 128 neurons, with dropout, LeakyReLU, and batch normalization applied after each layer. The second model provided a further baseline for a more complex (compared to the first proposed MLP) model, with a relatively higher learnable parameter count ($\sim$61k). All MLP models' output layers are linear projections to 3 units, representing the abundances. Despite its lack of complex model layers, the MLP provides a point of comparison for how well a non-spatial model can learn the unmixing task.

\subsubsection{Convolutional Neural Network:}\label{sec:Convolutional Neural Network}

The CNN treats each input spectrum as a one-channel 1D sequence and processes it through repeated convolutional stages, optionally with residual refinement with batch normalization in the reported configurations, GELU activation, and dropout in both the convolutional blocks and the output head, with optional residual blocks that add skip connections around pairs of convolutional blocks. Each stage ends with average pooling followed by adaptive average pooling and a small MLP head that outputs abundance fractions. CNN-M uses 3 convolution stages (64–256 filters, kernel size 7) and one residual block per stage. CNN-L increases depth by using two residual blocks per stage and width (96–384 filters), with 8\% dropout. CNN-XL adds a fourth convolutional stage, wider filters (up to 512), larger kernels (size 9), and 10\% dropout. Residual blocks are enabled across all reported CNN variants.

\subsubsection{Fourier Neural Operator:}\label{sec:Fourier Neural Operator}

The Fourier Neural Operator (FNO) model employs global convolution via Fourier transforms to capture long-range spectral patterns ~\cite{li2020}. Given that spectra have long-range spectral patterns themselves due to the emission bands of different materials, we chose FNO as a candidate. It treats the input as a 1D signal and performs spectral convolution in the frequency domain. The medium FNO (FNO-M) consists of 4 Fourier blocks with a hidden width of 128 and uses the lowest 32 frequency modes in each layer. The large FNO (FNO-L) maintains 4 blocks but expands to a width of 192 and 48 Fourier modes. The extra-large FNO (FNO-XL) uses 5 blocks, 256 feature width, and 64 modes. All FNO variants apply a global mean pooling before the output layer (no local pooling between layers) and do not use dropout. Like the CNN, the FNO's final layer is a linear projection to the 3 abundance values.

\section{Ground-Truth Dataset Results}\label{sec:Ground-Truth Dataset Results}

We begin by evaluating how well different architectures perform spectral unmixing when trained on ground-truth data. This serves as a baseline: our ultimate goal is to compare these results with models trained on synthesized data to assess whether a learned synthesizer can replace or augment scarce ground-truth labels, and to use this baseline to choose which unmixing models to test further.

All models were trained on the dataset under comparable settings (using the same train, validation, and testing split and up to 80 epochs of training with early stopping). The $R^2$ values were the primary metric of interest, as they provide a relevant statistical measure of how well an unmixing model performs.

The FNO models achieved the highest overall accuracy, as indicated by their high total $R^2$ values. FNO-L and FNO-XL achieved the lowest test MSE ($\sim$0.0084–0.0087) and the highest total $R^2$ ($\sim$0.91), outperforming both CNN and MLP variants. A clear scaling trend appears: increasing model size from FNO-M to FNO-L notably improved accuracy, while FNO-XL offered smaller additional gains. This suggests that FNO benefits from larger capacity, though with diminishing returns at the XL scale.
The CNN models showed the opposite trend: CNN-M achieved the best test total $R^2$ ($\sim$0.86), while larger models like CNN-L and CNN-XL performed worse, with CNN-XL dropping to ~0.79. This suggests that increasing CNN capacity led to only slightly greater overfitting rather than improved generalization. All CNNs were still improving near the end of training (best epochs around 75–76 out of 80), thus larger models may have benefited from further tuning.

The MLP-S baseline reached a test total $R^2$ of $\sim$0.88, comparable to CNN-M and close to the FNO models. Additionally, the MLP-M baseline model reached a test total $R^2$ of $\sim$0.89. Despite lacking explicit spectral convolution, the MLP models performed competitively on the per-pixel unmixing task, suggesting that much of the task can be learned without it. However, the FNO's ability to capture more global spectral structure gave it an edge in accuracy.

Across all models, we observed minimal overfitting. The generalization gaps were very small (on the order of $10^{-3}$), indicating that validation and training errors remained close. Even the largest models (CNN-XL, FNO-XL) showed only a slight increase in gap ($\sim0.005-0.007$ MSE), and in earlier 80-epoch experiments, some gaps were essentially zero or negative. These tiny gaps suggest no severe overfitting, suggesting that the models may have benefited from additional training. In summary, the FNO architecture delivered the best unmixing performance in our experiments, while the CNN and MLP provided strong baselines. The results also highlight that model scaling can have different effects: FNO gains from scaling (up to saturation), whereas the CNN requires further tuning or longer training to benefit from increased capacity.

Given these results, we selected FNO-XL and MLP-M as the primary unmixing models for evaluating synthesis-based training, as they have the highest overall accuracy, and training will be conducted on dedicated hardware. The FNO-XL model was selected primarily because of its high complexity and its ability to achieve the highest total $R^2$ values in testing. Additionally, MLP-M was chosen for achieving reasonable total $R^2$ values and lower complexity. The disparity in complexity between the two models was deliberately chosen to test one of the assumptions of the proposed decoupling framework, as stated in Section \ref{sec:Summary of Assumptions and Claims}.

\section{Generated Dataset Results}\label{sec:Generated Dataset Results}

Presented in this section are tests conducted on generated datasets using the Gaussian Diffusion, based on a 1D Conditional U-Net with Conformer Layers and a Dual-Path Transformer Conditional Variational AutoEncoder, named ``GD-Streamline" and ``TCVAE," respectively.

\subsection{Manifold Approximation}\label{sec:Manifold Approximation}

\begin{table}[h]
	\centering
		\begin{tabular}{|l|c|c|c|c|}\hline
			Model & Distance Type & $\delta$ & $\zeta$ & $\epsilon$ \\ \hline
			GD-Streamline & Euclidean & 4.0682 & -10 & -11 \\
			  GD-Streamline & Spectral Angle & 4.0681 & -6 & -7 \\
			  TCVAE & Euclidean & 5.5094 & -50 & -49 \\
            TCVAE & Spectral Angle & 14.000 & -24 & -23 \\ \hline
		\end{tabular}
	\caption{Comparison of nearest neighborhood metrics across different trained generators.}
\label{table:manifold_approx}
\end{table}

As presented in Table \ref{table:manifold_approx}, the Gaussian Diffusion model, GD-Streamline, failed dramatically in approximating the manifold of the ground-truth dataset; the $\delta$ values were relatively high for both Euclidean and spectral angle distance definitions, with values 4.0682 and 4.0681, respectively. Additionally, the $\zeta$ metric is observed to be negative at -10 and -6, indicating considerable model hallucination. Finally, a negative $\epsilon$ indicates higher clumping in the generated spectra than in the ground-truth spectra for both distance definitions, with values of -11 and -7, respectively. 

The manifold approximation capability of the TCVAE model, as shown by the nearest neighbor metrics presented in Table \ref{table:manifold_approx}, indicates a domain collapse worse than GD-Streamline. With the $\delta$ values of 5.5094 and 14.000 for the Euclidean and spectral angle distance definitions, respectively, the domain collapse is comparatively worse than that with GD-Streamline. Additionally, the generated spectra exhibit more clumping than the ground-truth spectra under both distance definitions, as indicated by $\epsilon$. Additionally, the TCVAE model showed a greater hallucination in spectral amplitudes than in spectral shapes, as measured by the spectral angle, as indicated by the $\zeta$ metric.

\subsection{Effects of Decoupling on Unmixing Accuracy}\label{sec:Effects of Decoupling on Unmixing Accuracy}

\begin{table}[h]
	\centering
		\begin{tabular}{|l|c|c|c|c|}\hline
			$G$ & $U$ & DS & $R^2$-T & $R^2$-NPV \\ \hline
			GD-St. & FNO & 3k & -1.672 & -1.996 \\
			  GD-St. & FNO & 9k & -1.644 & -1.986 \\
            GD-St. & FNO & 27k & -1.672 & -1.968 \\
			  GD-St. & MLP & 3k & -1.589 & -1.987 \\
            GD-St. & MLP & 9k & -1.766 & -1.984 \\
			  GD-St. & MLP & 27k & -1.558 & -1.884 \\
			  TCVAE & FNO & 3k & -2.072 & -4.531 \\
            TCVAE & FNO & 9k & -1.133 & -2.248 \\
            TCVAE & FNO & 27k & -1.065 & -2.216 \\
            TCVAE & MLP & 3k & 0.434 & 0.065 \\
            TCVAE & MLP & 9k & 0.352 & -0.058 \\
            TCVAE & MLP & 27k & 0.307 & -0.140 \\ \hline
		\end{tabular}
	\caption{Effect of decoupling, under different generators, unmixing models, and generated dataset sizes.}
\label{table:decoupling-effect}
\end{table}

The unmixing accuracies during testing on a sub-section of the ground-truth dataset used during the generator's validation and testing sequences are presented in Table \ref{table:decoupling-effect} for an unmixing model trained on data generated by the trained generator. Unmixing accuracies were assessed using the coefficient of determination ($R^2$) relative to actual and predicted abundances during model testing. Both mean $R^2$ values of different endmember classes averaged over, and NPV-only $R^2$ values were presented. An ablation study across 2 different trained generator models, 2 different unmixing architectures, and 3 different generated dataset sizes in terms of the amount of spectra, denoted as ``$G$", ``$U$", and ``DS", is presented in Table \ref{table:decoupling-effect}.

The unmixing models trained on datasets generated by the primary Gaussian Diffusion model, GD-Streamline, underperformed during testing, with $R^2$ values consistently negative. Additionally, no distinct correlation between dataset size and the marginal change in $R^2$ accuracy is apparent, as shown in Table \ref{table:decoupling-effect}. The unmixing models' accuracies in predicting the NPV abundance class have systematically underperformed those for predicting any endmember class, on average.

\section{Discussion}\label{sec:Analysis and Discussion}

\begin{figure}[htp]
    \centering
    \includegraphics[width=\columnwidth]{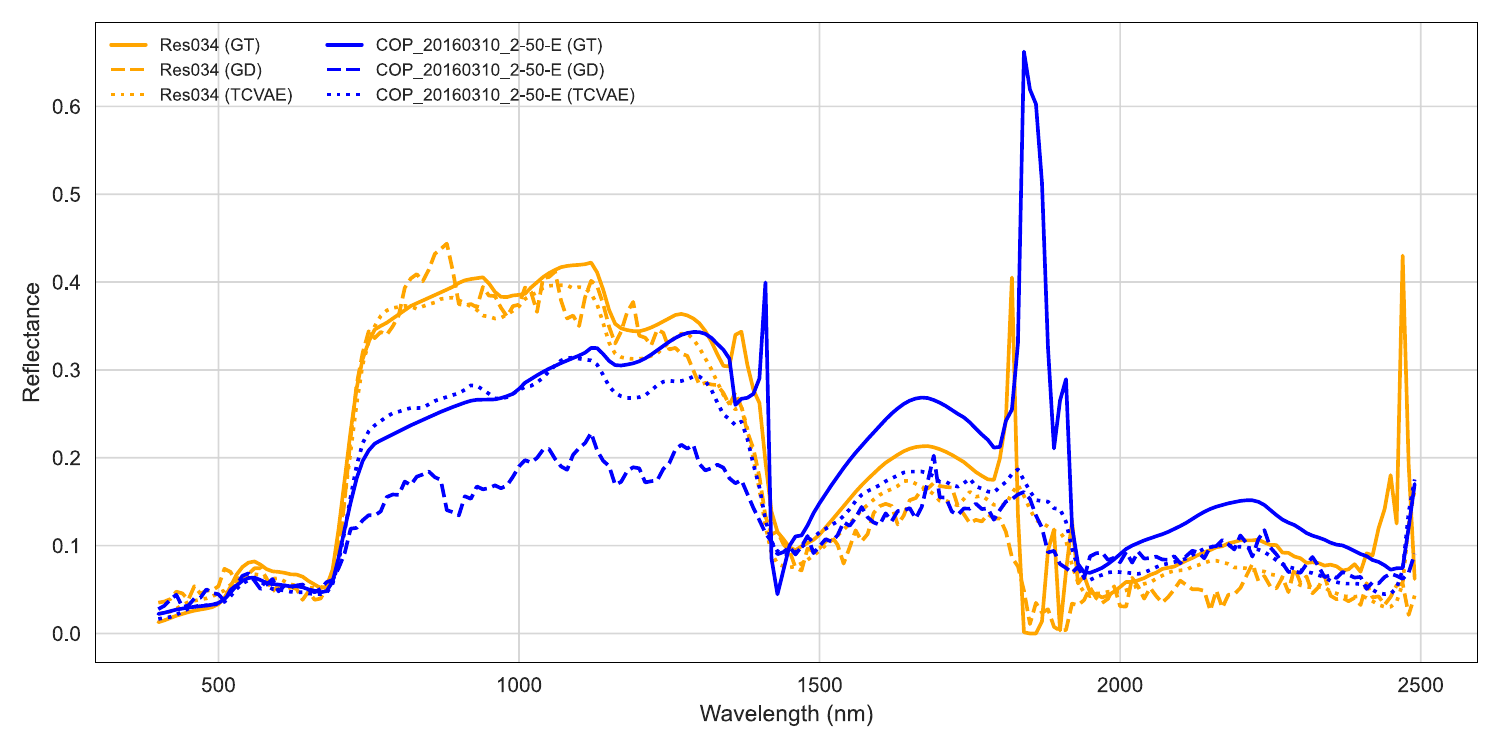}
    \caption{Visual Heuristics of Ground-Truth and Generated Spectra for Two Representative Abundance Conditions Across GD-Streamline and TCVAE}
    \label{fig:Sanity Check}
\end{figure} 

Figure \ref{fig:Sanity Check} indicates, to an extent, a successful approximation of overall spectral features. Consequently, the suboptimal testing metrics presented in Sections \ref{sec:Manifold Approximation} and \ref{sec:Effects of Decoupling on Unmixing Accuracy} indicate a significantly high sensitivity to spectral features in approximating the underlying data manifold.

\subsection{Manifold Approximation and Domain Collapse}\label{sec:Manifold Approximation and Domain Collapse}

The Gaussian Diffusion Streamline (GD-Streamline) model consistently underperformed in both manifold approximation and, consequently, performance yield within the decoupling framework. Although heuristically the GD-Streamline models produce noisy spectra that resemble the ground-truth dataset spectra, the $\delta$-nearest-neighbor metric strongly indicates severe domain collapse, failing to approximate the ground-truth dataset's underlying manifold. Across different nearest-neighbor distance definitions, the metrics did not change drastically, indicating a severe domain collapse in both spectral shape features and amplitudes. Such a collapse indicates that the TV, SAM, and FFT physical priors did not provide a significant benefit beyond heuristic improvements. As indicated by the $\zeta$ metric, the GD-Streamline also hallucinates to some extent; with a negative $\epsilon$ indicating a form of mode collapse, in which generated spectra tend to be more clumped than in the ground-truth dataset. 

As stated in Section \ref{sec:Manifold Approximation}, the TCVAE model underperformed in manifold approximation, comparatively worse than the GD-Streamline model. With the disparity in $\delta$ metrics defining domain collapse, which changes dramatically with different distance definitions, indicating that the TCVAE model lacked the necessary physical priors to correctly approximate the spectral manifold underlying their features. Additionally, with the comparatively high $\epsilon$ metric, it is evident that the TCVAE model experienced severe domain collapse, resulting in most spectra collapsing into a specific spectral region and possibly leading to hallucinations, as shown by the $\zeta$ metric. Although the TCVAE model presented superior heuristics compared to GD-Streamline, it was plagued by small-scale noise, as indicated by nearest-neighbor metrics, and its manifold approximation was comparatively worse.

\subsection{Unmixing Performance and Accuracy Yield of Decoupling}\label{sec:Unmixing Performance and Accuracy Yield of Decoupling}

Given the severe implications of the nearest-neighbor metrics applied to GD-Streamline, the decoupling yields, as expected, severe unmixing inaccuracies. With the implication of high $\delta$, that the learned physics and underlying manifold are different from the ground-truth, the unmixing models have consequently created a mapping between nonphysical, or at best, wrongly approximated manifolds. Given the severe domain collapse, the similarity in failure magnitudes between the FNO and MLP unmixing models was expected.

In contrast to the comparatively worse manifold approximation metrics, the TCVAE model achieved much higher accuracy in decoupling than the GD-Streamline model. Although the TCVAE model showed worse domain collapse than GD-Streamline, the much higher accuracy achieved with decoupling suggests the possibility of localized mode collapse. Given the significantly high clumping of generated spectra, the domain collapse, hallucination, low NPV $R^2$ accuracy, yet relatively high Total $R^2$ accuracy, it is plausible that the TCVAE model has learned and domain collapsed on a narrow section of the ground-truth manifold $\mathcal{M}_n$. Additionally, the disparity in the increasing size of the generated dataset's effect on FNO and MLP model predictions, with a positive and negative effect, respectively, indicates that the high-capacity unmixing model requirement stated in Section \ref{sec:Summary of Assumptions and Claims} can be met to an extent by an FNO architecture. 

\subsection{Limitations of the Study}\label{sec:Limitations of the Study}

\subsubsection{Computational Limitations:}

The scope of this study was inherently constrained by the computational resources available. Such a limitation naturally constrained the exploration of the scaling of the two generative model architectures, in terms of the total number of learnable parameters and the number of trained epochs. As a result, the models were naturally not scaled to an extent in terms of learnable parameters and training epochs to reasonably observe the effect of double-descent ~\cite{nakkiran2019}, which through the asymptotic reduction of training losses for both $U$ and $G$ models, can in theory reduce both $\epsilon_{app}(U)$ and $\epsilon_{gen}$. Given that reaching the second-descent regime for deep generative architectures such as those employed in this study is often critical for generalization ~\cite{fraij2025}, it is theoretically possible that the models in this study lacked the necessary capacity or training duration to reach it. Additionally, due to limited computational resources, observing the asymptotic behavior of the dataset size ($k$) as it increased was constrained. 

\subsubsection{Ground-Truth Dataset Heterogeneity:}

Furthermore, the ground-truth dataset utilized throughout the study for training, validation, and testing presents significant challenges due to its inherent heterogeneity. The dataset, an aggregation of six individual studies, exhibits inherently different mixing physics and abundance distributions, suggesting the possibility of a highly discontinuous or volatile underlying ground-truth dataset manifold $\mathcal{M}_n$. Covariate shift and distributional mismatches, given the dataset heterogeneity, could have theoretically increased both bias terms ($\epsilon_{app}(U)$ and $\epsilon_{gen}$) and the risk minimization term on the generated dataset ($\widehat{R}_k(U)$). A ground-truth dataset with less heterogeneity could, in theory, have yielded better manifold approximation metrics and a stronger decoupling effect.

\section{Conclusion}\label{sec:Conclusion}

This study investigated the effects of decoupling the interpolation and mapping tasks of a regular unmixing model on unmixing accuracy. It established the critical importance of including physical priors during the training of the generator model $G$. 

This was demonstrated by the comparatively superior manifold approximation performance of GD-Streamline, incorporating several different physical priors during training, relative to TCVAE. However, it was also concluded that physical priors introduced during the training of the generator models, such as TV and SAM, were not entirely effective at improving the unmixing accuracy of the unmixing model $U$ trained on generated data, and that more rigorous physical priors were needed. 

The mismatch in TCVAE's $R^2$ values indicates the difficulty of approximating the effects of NPV abundance condition in the ground-truth manifold. Conversely, comparisons with the GD-Streamline's $R^2$ values indicate that the KL-Divergence component in the TCVAE model was instrumental in approximating parts of the manifold. 

Finally, the scaling of the generated dataset size ($k$) is theoretically viable for decoupling, as in specific generator-unmixing model combinations (TCVAE and FNO, respectively), a positive correlation between dataset size and decoupling efficacy was observed.

\subsection{Future Work}\label{sec:Future Work}


\subsubsection{Non-Photosynthetic Vegetation Abundance Mapping:}\label{sec:Non-Photosynthetic Vegetation Abundance} 
Given the established difficulty in approximating the influence of NPV abundance on ground-truth manifolds, it is suggested that developing abundance mappings with greater expressivity with respect to class-wise differences may yield improvements in both generation and unmixing performance.

\subsubsection{End-To-End Hyperparameter Tuning:}\label{sec:End-To-End Hyperparameter Tuning}
Due to development constraints, the two generative models were developed independently of the unmixing pipeline. Consequently, the hyperparameters of both models were tuned using visual heuristics and model-specific metrics, such as the TCVAE's reconstruction loss. Given the sensitivity of model performance to hyperparameter selection, a promising direction for future work would be to optimize hyperparameters end-to-end against the metrics used in this study, rather than the two-stage development pipeline.

\subsubsection{Flow-Matching Models:}\label{sec:Flow-Matching Models}
In this work, we investigated two classes of generative models: diffusion models and conditional variational autoencoders, and our empirical results demonstrate that the choice of model class has a notable effect on the characteristics of the generated output. Specifically, we observed a smoothing phenomenon in the outputs of our TCVAE model, whereas our Gaussian Diffusion model exhibited elevated noise in its generated spectra. We hypothesize that the limited volume of training data amplifies each model's inductive bias in its generated output, as it must rely more heavily on its structural assumptions when the empirical signal is scarce. It would therefore be worthwhile to examine the generated outputs produced by flow-matching models, which offer a distinct set of inductive biases compared to the model classes studied here.

\section*{Acknowledgments}\label{sec:Acknowledgments}

We would like to extend our sincere gratitude and appreciation to the University of Toronto Aerospace Team's (UTAT) Space Systems Division for providing foundational support, a collaborative research environment, and the operational framework that were critical to conduct this study. Furthermore, we would like to thank our colleagues and teammates for their meticulous proofreading, which contributed significantly to the clarity and quality of this manuscript.

{
	\begin{spacing}{1.17}
		\normalsize
		\bibliography{refs} 
	\end{spacing}
}

\section*{Appendix}\label{sec:Appendix}

The code, configuration files, and images used in this study can be found in the following GitHub Repository Release: UTAT FINCH Mission Synthetic Data Generation (v1.0.0). The frozen code used for this manuscript is persistently accessible via Zenodo in: \url{https://doi.org/10.5281/zenodo.19324459}. For the actively maintained codebase and future updates, please refer to the repository: \url{https://github.com/utat-ss/FINCH-Science_SyntheticData}.

\end{document}